\documentclass[10pt,twocolumn,letterpaper]{article}

\usepackage{cvpr}
\usepackage{times}
\usepackage{epsfig}
\usepackage{graphicx}
\usepackage{amsmath}
\usepackage{amssymb}

\usepackage{xcolor}

\usepackage[pagebackref=true,breaklinks=true,letterpaper=true,colorlinks,bookmarks=false]{hyperref}

\cvprfinalcopy 

\def\cvprPaperID{6} 

\ifcvprfinal\fi
\begin{document}

\title{Semantic Binary Segmentation using Convolutional Networks without Decoders}

\author{Shubhra Aich, William van der Kamp, Ian Stavness\\
Department of Computer Science\\University of Saskatchewan, Canada\\
{\tt\small \{s.aich, william.vanderkamp, ian.stavness\}@usask.ca}
}

\maketitle

\begin{abstract}

In this paper, we propose an efficient architecture for semantic image segmentation using the depth-to-space (D2S) operation. Our D2S model is comprised of a standard CNN encoder followed by a depth-to-space reordering of the final convolutional feature maps. Our approach eliminates the decoder portion of traditional encoder-decoder segmentation models and reduces the amount of computation almost by half. As a participant of the \textit{DeepGlobe Road Extraction} competition, we evaluate our models on the corresponding road segmentation dataset. Our highly efficient D2S models exhibit comparable performance to standard segmentation models with much lower computational cost.

\end{abstract}

\section{Introduction}

\begin{figure*}[t]
\centering
	\includegraphics[scale=0.32]{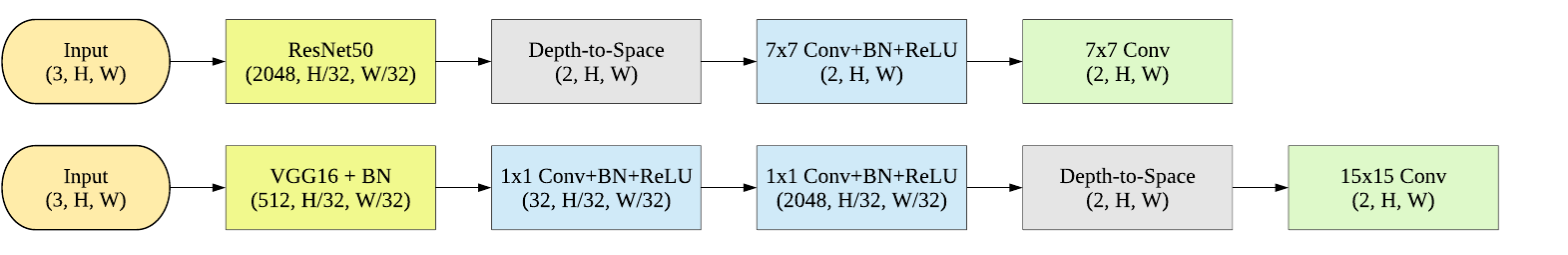}
    \caption{D2S models with ResNet50 (top) and VGG16-BN (bottom) backbones. Because of the differences in the shape of the final output layer, the placement of the rearrangement or depth-to-space block is different for these models. The last one or two convolution operations incur a negligible computational cost due to the small number of channels (2).}
    \label{fig:architecture}
\end{figure*}

Semantic segmentation refers to classifying the pixels of images or videos according to specific categories of objects or background regions known as stuff \cite{coco-stuff}. Like many other areas of computer vision, research on semantic segmentation has received a tremendous performance boost with the emergence of deep learning in recent years. All recent semantic segmentation models follow a general encoder-decoder type of architecture where the encoder front-end of the network extracts the features necessary for a particular task, and the decoder back-end of the network approximates the segmentation map from these salient features.

FCN \cite{fcn} is the earliest example of an encoder-decoder style semantic segmentation network. This architecture is built by converting the fully connected (FC) layers at the backend of traditional image classification architectures like AlexNet \cite{alexnet} or VGG \cite{vgg} into fully convolutional layers with $1\times 1$ convolution followed by upsampled or fractional convolution or deconvolution to generate the pixel-level segmentation map. Skip connections from the higher resolution layers at the convolutional front-end are added for better information gain or performance in both FCN and U-Net \cite{u-net}. Next comes the SegNet \cite{segnet} or deconvolutional network \cite{deconv} architectures, where for upsampling, max-pooling indices are used with the stack of simple convolutional layers. In SegNet, FC layers or equivalent convolution layers are omitted in order to reduce both the memory requirements and computational complexity of the network. Our network design has some similarities to both FCN and SegNet. First, like SegNet, we do not use any FC layers or their equivalent. Moreover, our network uses $1\times 1$ convolution like FCN, but with a much smaller size. Unlike FCN or SegNet, we do not use any deconvolution operation, rather a rearrangement of the feature grid is done by a depth-to-space operation with negligible computational cost.

Recent state-of-the-art segmentation approaches follow the traditional approach of requiring some sort of decoder back-end.
The DeepLab models \cite{deeplab-v1, deeplab-v2} use atrous convolutions \cite{atrous} in the backend of the CNN models to generate comparatively higher resolution coarse score maps ($1/8^{th}$ of the original image) instead of using max-pooling. They also use spatial pyramid pooling \cite{sppnet} with the atrous convolution of variable rates for better multi-scale prediction as well as fully connected CRF \cite{fc-crf} to finetune the bilinearly upsampled score maps.
Overall this results in a complex, multi-stage pipeline where CNN and CRF are trained separately, though the latest version of the model\cite{deeplab-v3} omits CRF post-processing.
%
RefineNet \cite{refinenet} uses a multi-path refinement architecture as its decoder. Each refinement block fuses high-level, and low-level feature maps using residual convolution layers and bilinear upsampling for shape adjustment. Also, the authors use the residual sequence of pooling for efficient fusion of multi-scale, pooled prediction.
The pyramid scene parsing network (PSPNet) \cite{pspnet} uses pyramid pooling on the feature map of the ResNet equipped with dilated or atrous convolution for global context aggregation. The authors also added a branch in the middle of the ResNet to propagate auxiliary loss for faster convergence.
The large kernel paper \cite{large-kernel-matters} uses larger convolution kernels to empirically cover larger receptive field \cite{zhou-detectors}. Large symmetric kernels are broken down into a few asymmetric kernels to reduce the computational complexity.
Finally, recent literature focused on road mapping from aerial images combine the power of deep learning for pixel-level segmentation with graph-based optimization for the extraction of road topology \cite{deep-road-mapper,roadmap-gan}.
All of these related works on semantic segmentation share the common feature of including a decoder sub-network composed of different variations of convolutional and/or upsampling blocks.

In this paper, we challenge the basic assumption that a decoder sub-network is needed to approximate a segmentation map from encoder-generated feature maps. We hypothesize that, at least for relatively easy segmentation tasks, such as binary segmentation, the computationally-complex decoder procedure can be replaced by a simple depth-to-space rearrangement of the output of the final convolution layer, without loss of segmentation accuracy. We call this type of encoder with depth-to-space (D2S) spatial reordering the D2S network. From an efficiency perspective, our D2S architecture needs only half of the computation to learn the same task.

The idea of depth-to-space reordering that we use in our paper to replace long-range decoders is identical to the sub-pixel convolution for image super-resolution \cite{depth-to-space}. Depth-to-space operations have also been used before for benchmarking different decoding approaches \cite{devil-decoder}, but in a different way. In that work, multiple instances of the depth-to-space reordering operation are used for $2\times 2$ upsampling in between the convolution layers in the decoder, whereas in our D2S model we use a single depth-to-space block as a replacement for a large stack of convolution layers.


We incorporate our D2S idea as a participant in the \textit{DeepGlobe Road Extraction} challenge. For the competition, our focus is to use a novel and efficient approach instead of an ensemble of sophisticated models. We evaluate our approach on the road extraction dataset. Our D2S model based on the ResNet50 encoder achieves 60.60\% mean intersection over union (IoU) whereas the top entry has IoU of 65.60\% on the validation set (at the time of writing). This small difference with the best entry validates our hypothesis that for at least easy segmentation problems, encoder-only architectures without any decoder might be a reasonable and efficient model of choice.

\begin{figure*}[t]
\centering
	\includegraphics[scale=0.117]{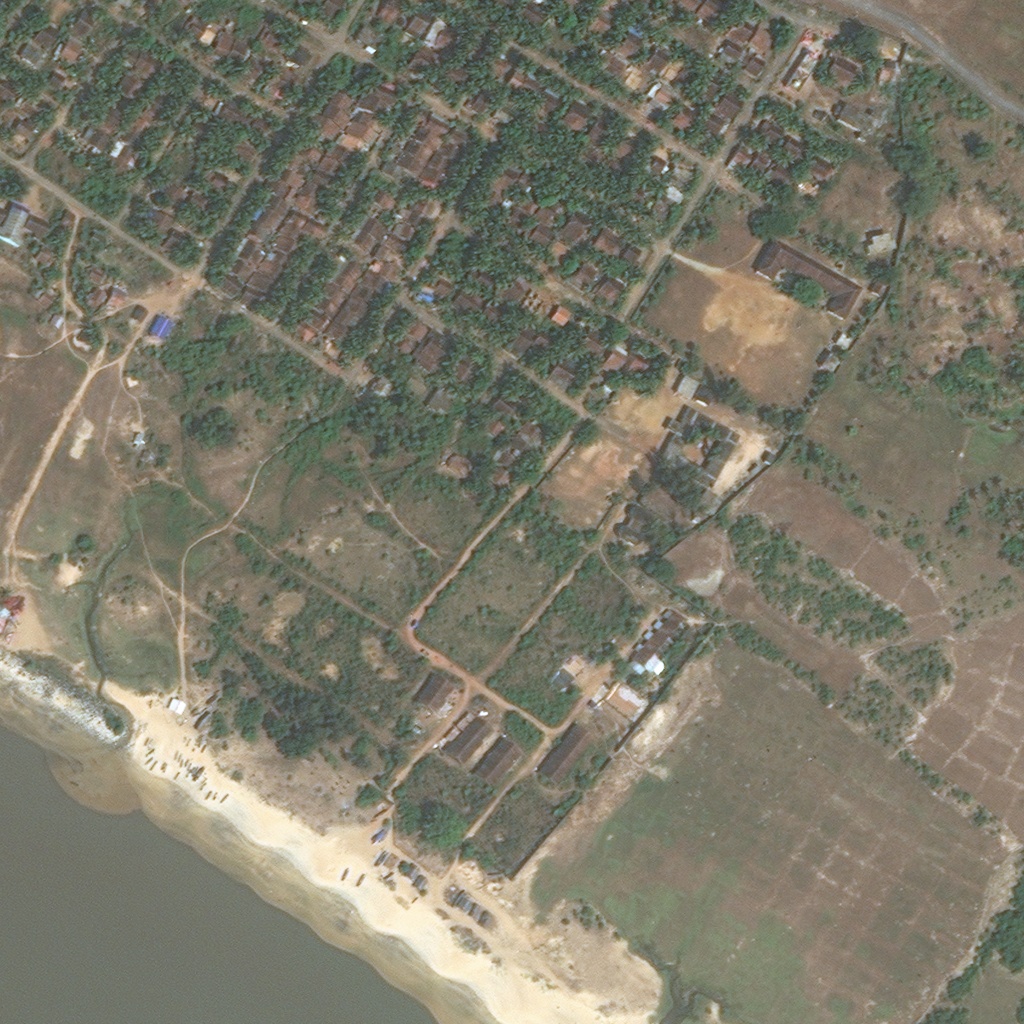}
	\includegraphics[scale=0.117]{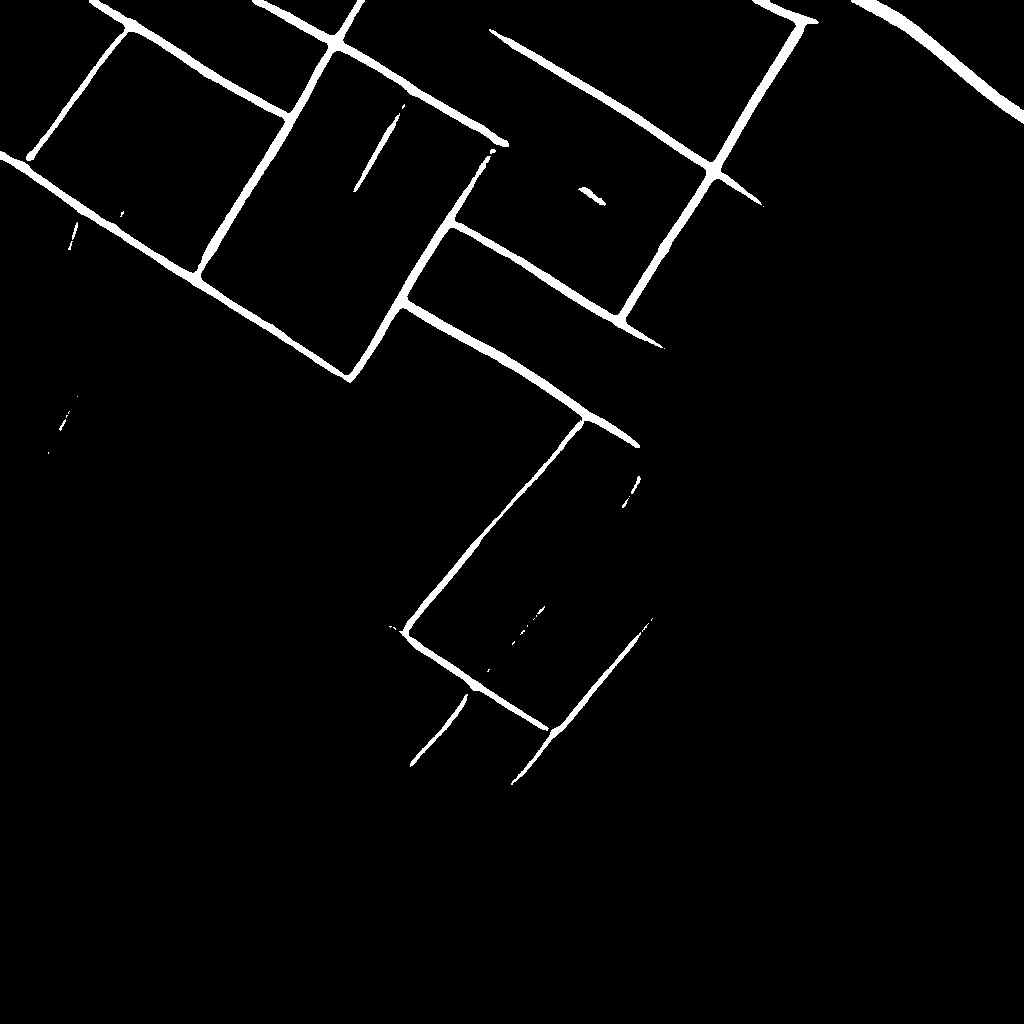}
	\includegraphics[scale=0.117]{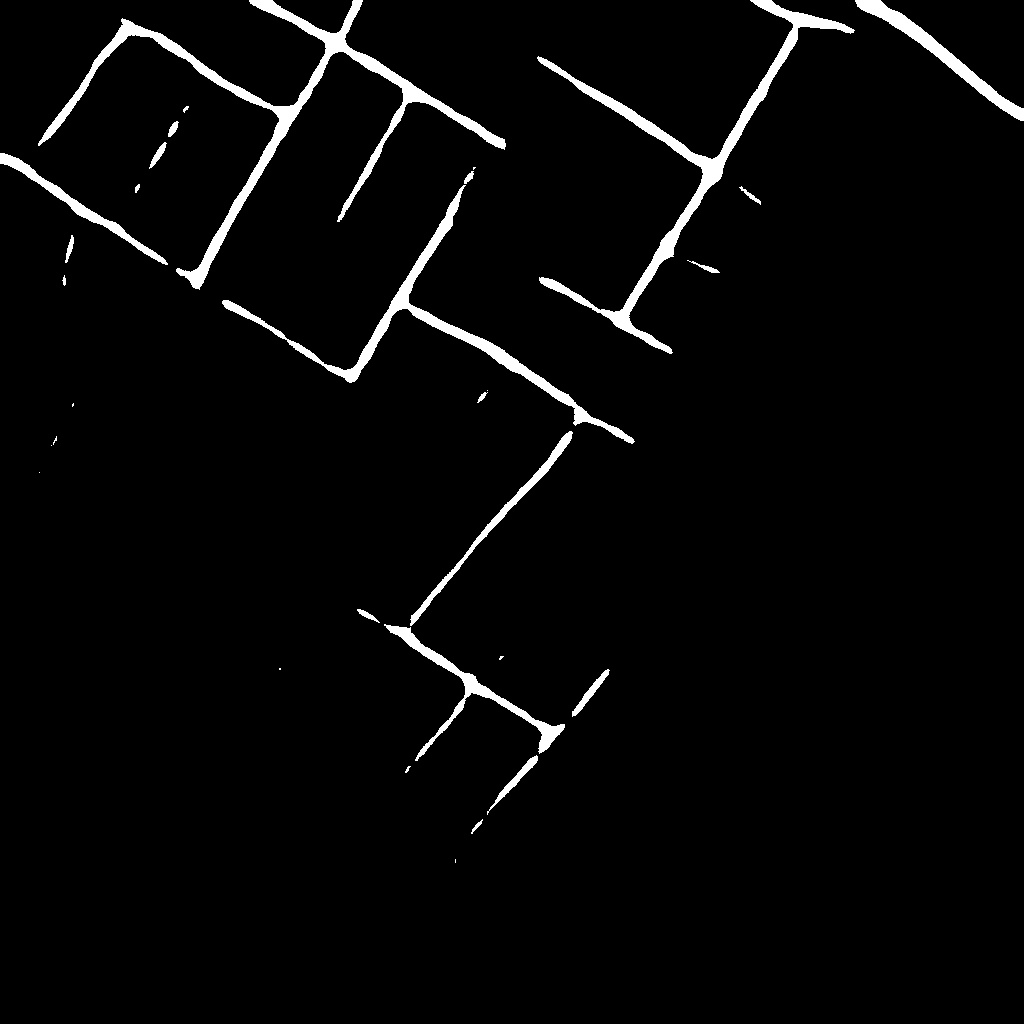}
	\includegraphics[scale=0.117]{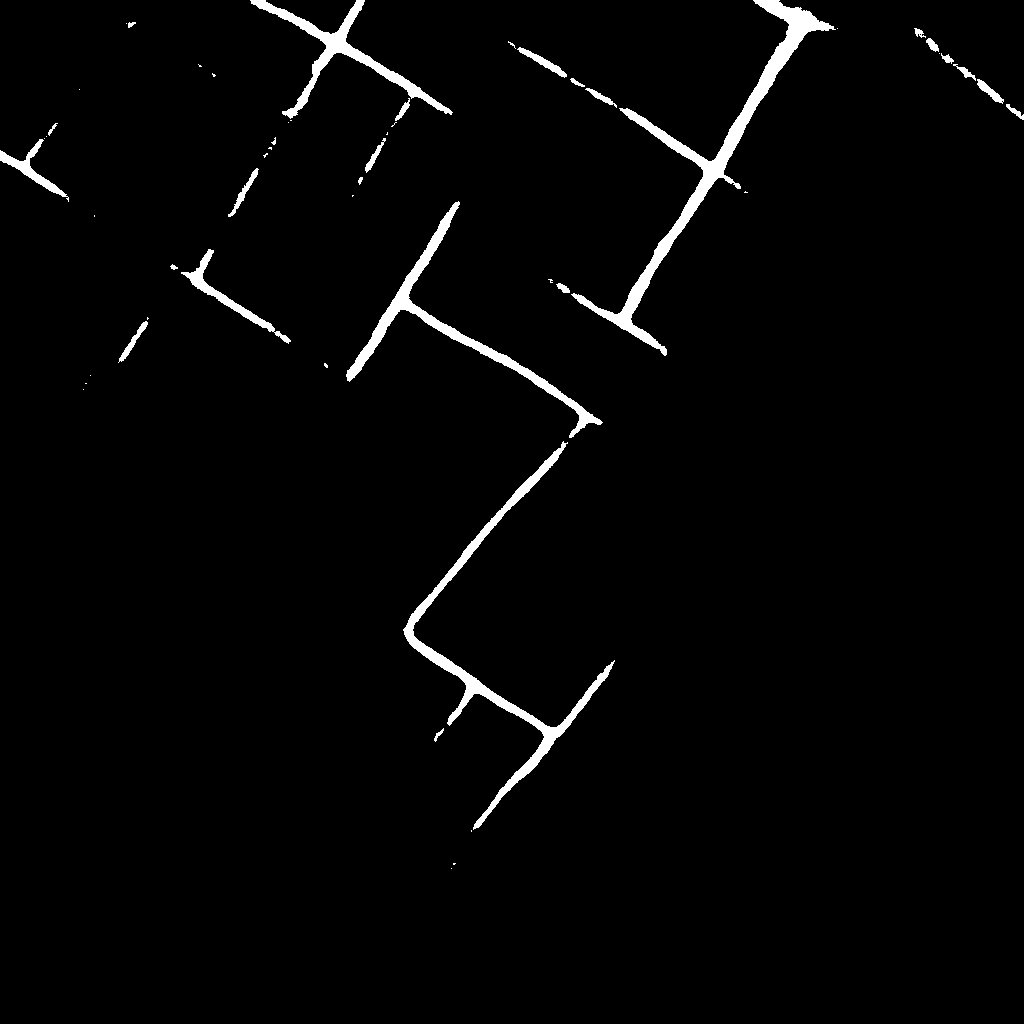} \\
	\includegraphics[scale=0.117]{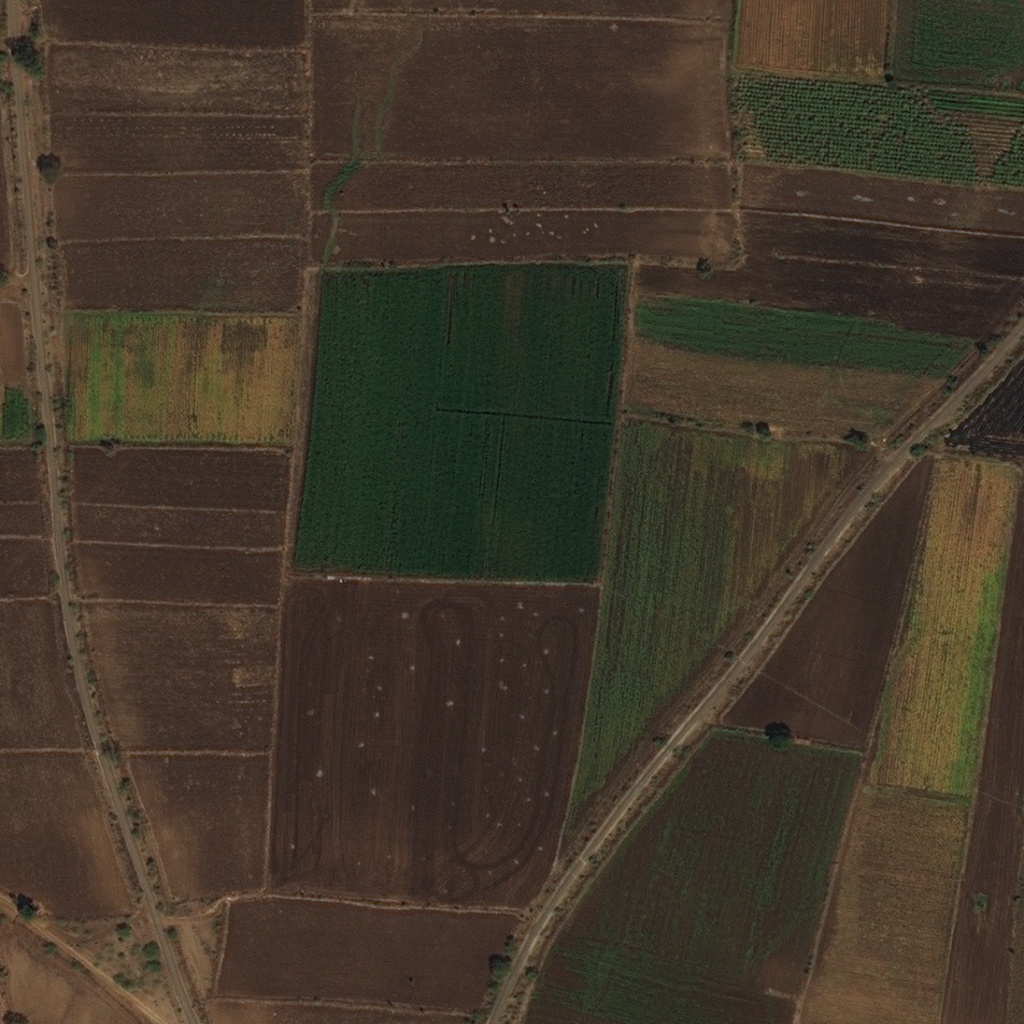}
	\includegraphics[scale=0.117]{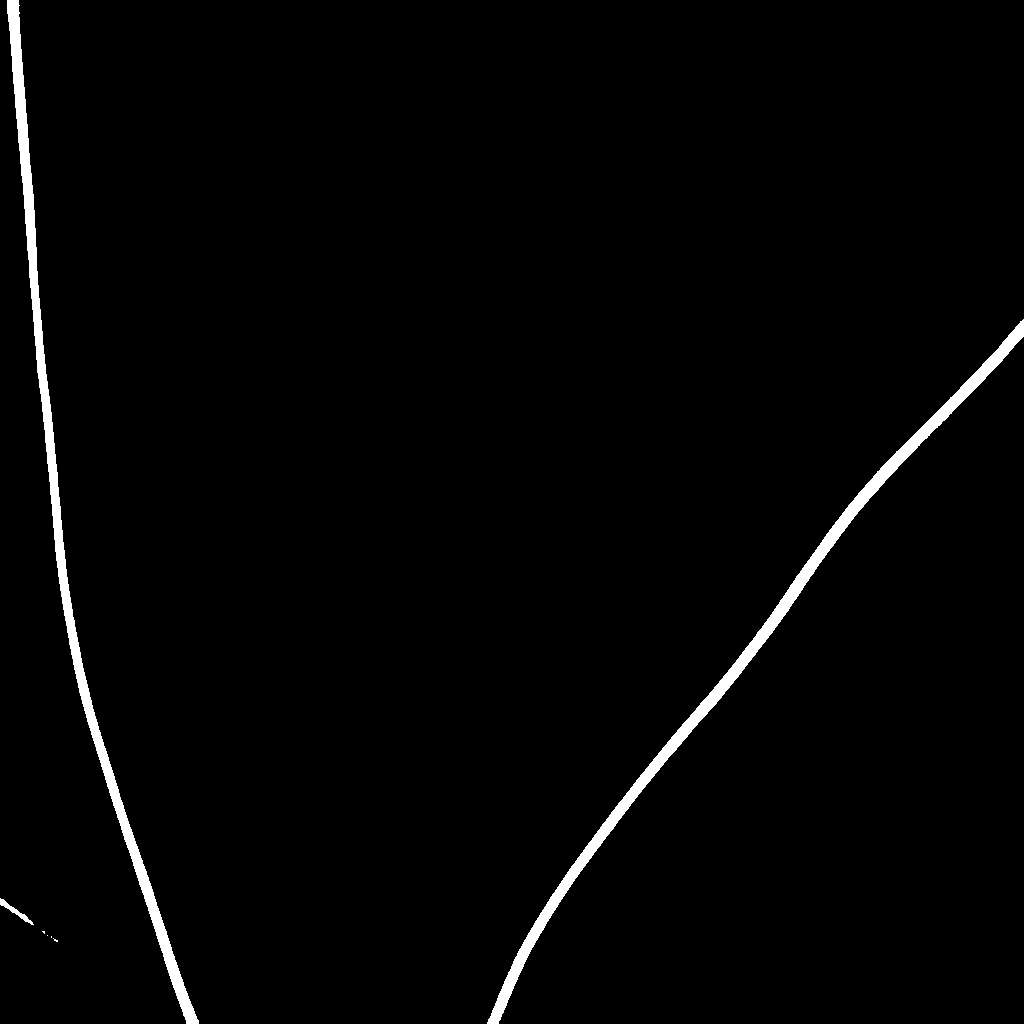}
	\includegraphics[scale=0.117]{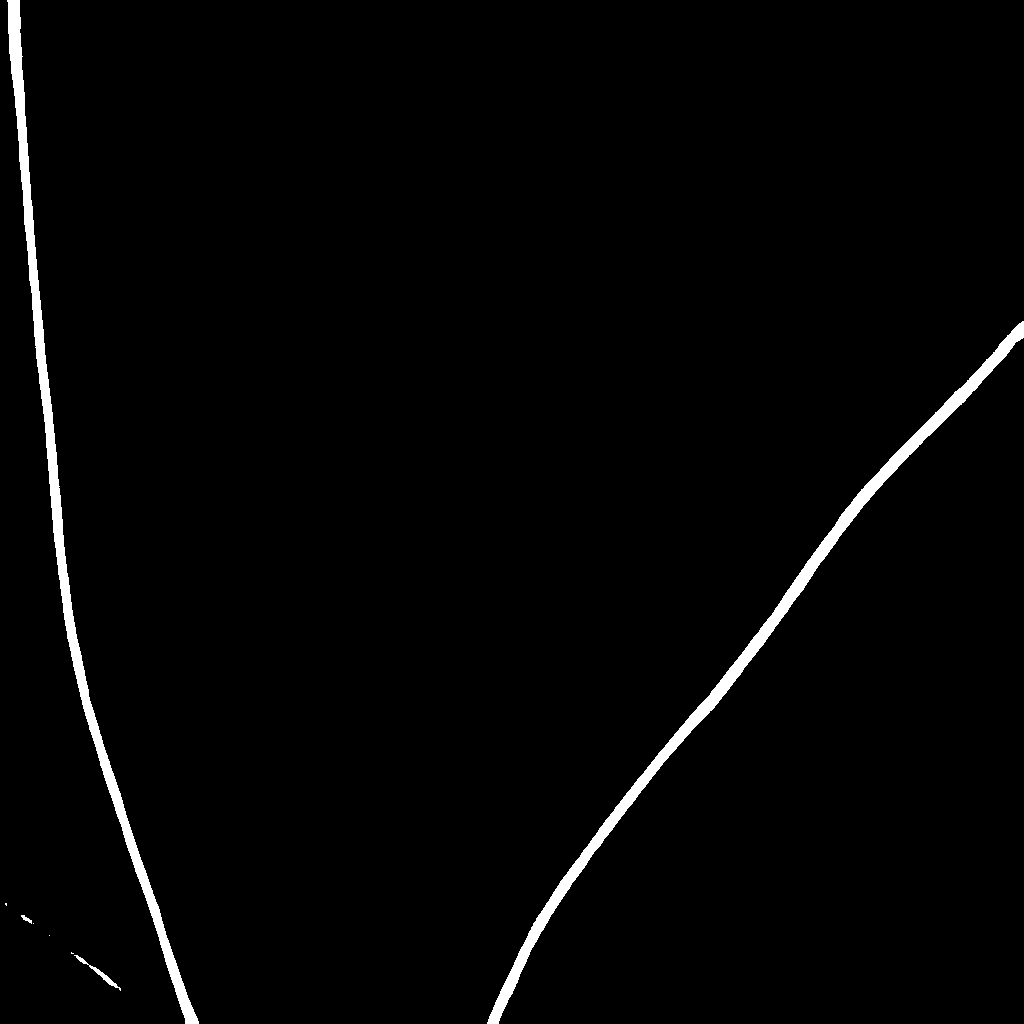}
	\includegraphics[scale=0.117]{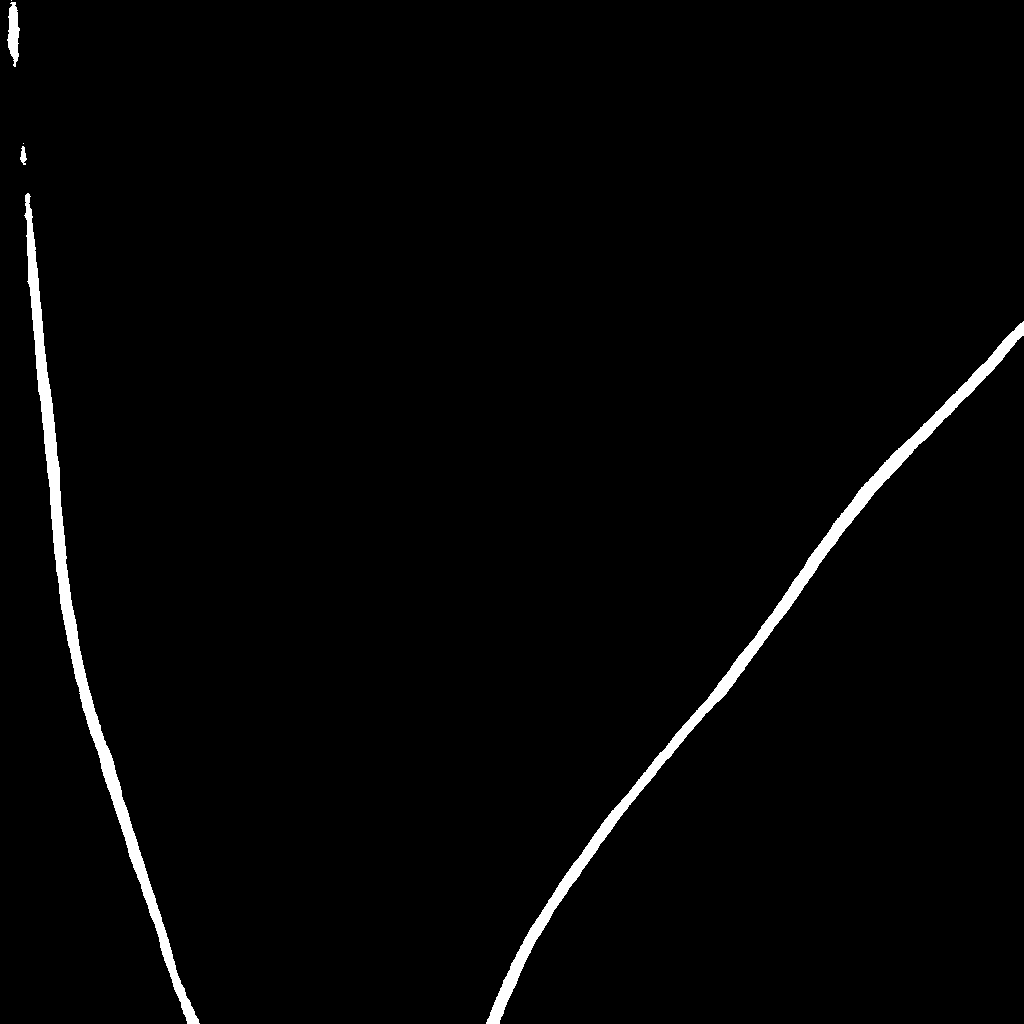}
    \caption{(Left to Right) Sample image; Segmentation maps generated by ResNet50-D2S, VGG16-BN-D2S, and Segnet models, respectively.}
    \label{fig:sample_images}
\end{figure*}

\section{Method}

With most current segmentation networks, such as FCN, DeepLab, or RefineNet, the predicted score map has a lower resolution than the input image. For SegNet, the shape of the input image and output segmentation map are the same with the same amount of computation in both encoder and decoder. For that reason, SegNet incurs twice the computational cost of the encoder alone.

In comparison, our architectural design allows pixel-wise prediction naturally. We train the network in such an arrangement where the neighborhood pixel contributions are stored along the depth dimension and then just reordered. One apparent drawback might be that the model will have artifacts in the final prediction map because the contextual mapping task in the neighborhood of the prediction map is interrupted. However, we did not see any such problems in practice, likely because the network learns to overcome the spatial disruption while training end-to-end.

\subsection{Architecture}

We employ Resnet50 \cite{resnet} and VGG16 \cite{vgg} with batch normalization \cite{batch-normalization} as the backbone or encoder of our network with minor differences due to the difference in the dimension of the output of the final convolution layer in these models. The complete architectures for both versions are depicted in Figure \ref{fig:architecture}.

We use a similar D2S reordering as proposed previously for the image super-resolution problem\cite{depth-to-space}. The domain of image super-resolution is a mapping task from the image space to itself with greater detail in the output space. Also, there is no encoder-decoder type of architectures; rather the raw image is taken as a dense feature map and a simple stack of convolution layers are used to produce the corresponding high-resolution version. Thus, the D2S transformation for super-resolution is an arguably more natural operation compared to our case, where we use this block right at the end of the encoder sub-network. In that sense, although we use a similar encoder type of network, it works as a decoder directly from the image space. This decoding task refers to mapping the RGB image pixels into the binary pixel space considering a semi-global context. Theoretically, the amount of context covered depends on the depth of the network. Moreover, we incorporate two-dimensional dropout \cite{dropout2d} after each max-pooling for the VGG model and after each block except the last one for the ResNet model for improved performance. For the VGG based model, we use $1\times 1$ convolution to obtain the depth necessary for pixel-level mapping (Figure \ref{fig:architecture}, bottom).


\begin{table}[t]
\centering
\caption{Results of our D2S models compared to SegNet as a baseline on the validation set.}
\label{tab:results}
\begin{tabular}{lc}
\hline
\textbf{Model} & \multicolumn{1}{l}{\textbf{Pixel IoU}} \\ \hline
ResNet50-D2S & 0.6060 \\ \hline
VGG16-BN-D2S & 0.5897 \\ \hline
SegNet \cite{segnet} & 0.5612 \\ \hline
\end{tabular}
\end{table}

\section{Experiments}

In this section, we provide a brief description of the dataset, the implementation and training details of our models, and results achieved on the validation set in the leaderboard compared to SegNet as a baseline.

\subsection{Dataset}

At the time of writing, the DeepGlobe Road Extraction dataset is only open for the participants of the ``DeepGlobe Road Extraction'' challenge \cite{deepglobe}. The dataset consists of 6226 and 1243 training and validation images, respectively, each of resolution $1024\times 1024$. This dataset is a binary image segmentation problem, where the road pixels are marked as foreground and the rest of the objects and stuff are background. One of the challenges of this dataset is that it is highly imbalanced in terms of the number of pixels per class, i.e. roads are thin lines within the images and therefore occupy few pixels as compared to the background.

\subsection{Training and Implementation}

We train all of our models with ImageNet \cite{imagenet-dataset} pretrained encoders. In the beginning, we started training with $224\times 224$ patches extracted from around the true positives in the ground truth due to the scarcity of the foreground (road pixels) in the images. However, empirically we found that having the full context of the image, i.e., training with the whole image at once helped to improve the model's performance.

We use PyTorch \cite{pytorch} as the deep learning framework. All the models are finally trained with full resolution images, and their color jittered versions with the batch size varying in the range of $[3, 8]$. The models are trained on NVIDIA TITAN Xp GPUs and an NVIDIA Quadro P6000 workstation. We use the Adam optimizer \cite{adam} with an initial learning rate of 0.0001 which is later reduced based on the training statistics. Codes and pre-trained models are publicly available. \footnote{\textcolor{red!50!black}{\texttt{https://github.com/littleaich/deepglobe2018}}}


\subsection{Results}

Table \ref{tab:results} lists the pixel-level intersection over union (IoU) \cite{deepglobe} for three different models on the validation set in the competition leaderboard. We provide the performance metric for a standard SegNet architecture to benchmark our D2S models for a couple of reasons. First, the front-end of our VGG-D2S model is a replica of the SegNet encoder, which is the set of convolution layers of the VGG16 model with batch normalization. Therefore, it is more straightforward to compare the symmetric decoder of SegNet against our spatial rearrangement strategy. Second, SegNet has been reliably employed for binary image segmentation problems with substantial accuracy in recent works \cite{aich-cvppp, deepwheat}.

From Table \ref{tab:results}, we find the D2S models to have comparable performance to the SegNet architecture. Figure \ref{fig:sample_images} shows two sample images and their corresponding segmentation maps generated by the three models. From this figure, it is also evident the qualitative performance of the models are quite similar.

Moreover, at the time of writing, the top entry in the leaderboard had IoU of 0.6560 which is $\sim 5\%$ better than our best model. We anticipate that like other featured competitions, the top entries in this competition comprise an ensemble of different approaches, whereas our result is generated using only the D2S models described in this paper. Therefore, we conclude that for segmentation problems containing only a few classes, heavy-decoder models like SegNet can be reliably replaced by our efficient D2S architecture without significant loss in performance.

\section{Conclusion}

In this paper, we propose an efficient image segmentation network, called D2S, that uses only a convolutional encoder along with spatial reordering of the final feature maps. Empirically, we show that for relatively easier image segmentation problems, such as binary segmentation, the D2S models give comparable performance to the standard models. Although we only evaluate our model on a simpler problem, this kind of depth-to-space architecture may also be useful in more complex tasks, which we plan to explore in future research.

\section*{Acknowledgment}

This research was undertaken thanks in part to funding from the Canada First Research Excellence Fund, the Natural Sciences and Engineering Research Council (NSERC) of Canada, and the Microsoft AI for Earth and NVIDIA GPU programs.

{\small
\bibliographystyle{ieee}
\bibliography{egbib}
}

\end{document}